\documentclass[letterpaper,man,apacite,floatsintext]{apa7}
\usepackage{graphicx}
\usepackage{natbib}
\usepackage{framed,enumitem} 
\usepackage{upgreek}
\usepackage{siunitx}
\usepackage{adjustbox}
\usepackage{multirow}
\usepackage{longtable}

\usepackage{subfiles}
\usepackage{xr-hyper}
\usepackage{csquotes}
\usepackage{comment}
\newcommand{\sgt}{social group token}
\usepackage{subfigure}
\usepackage{amsmath}
\usepackage{amssymb}

\title{Hate Speech Classifiers Learn Human-Like Social Stereotypes}

\shorttitle{Social stereotypes and hate speech detection}
\authorsnames[{1,3}, {2,3}, {1,3}, {1,2,3}]{Aida Mostafazadeh Davani, Mohammad Atari, Brendan Kennedy, \\Morteza Dehghani}
\authorsaffiliations{{Department of Computer Science, University of Southern California}, {Department of Psychology, University of Southern California}, {Brain and Creativity Institute, University of Southern California}}

\authornote {Correspondence regarding this article should be addressed to Aida Mostafazadeh Davani, mostafaz@usc.edu, 362 S. McClintock Ave, Los Angeles, CA 90089-161.
This research was supported by National Science Foundation (CAREER BCS-1846531).}

\abstract{Social stereotypes negatively impact individuals' judgment about different groups and may have a critical role in how people understand language directed toward minority social groups. Here, we assess the role of social stereotypes in the automated detection of hateful language by examining the relation between individual annotator biases and erroneous classification of texts by hate speech classifiers. Specifically, in Study 1 we investigate the impact of novice annotators' stereotypes on their hate-speech-annotation behavior. In Study 2 we examine the effect of language-embedded stereotypes on expert annotators'  judgements in a large annotated corpus. Finally, in Study 3 we demonstrate how language-embedded stereotypes are associated with systematic prediction errors in a neural-network hate speech classifier. Our results demonstrate that hate speech classifiers learn human-like biases which can further perpetuate social inequalities when propagated at scale. This framework, combining social psychological and computational linguistic methods, provides insights into additional sources of bias in hate speech moderation, informing ongoing debates regarding fairness in machine learning.}

\begin{document}

\maketitle

Artificial Intelligence (AI) technologies are prone to acquiring cultural, social, and institutional biases from the real-world data on which they are trained \citep{mccradden2020patient, mayson2018bias, o2016weapons, mehrabi2019survey, leavy2018gender, obermeyer2019dissecting, chouldechova2017fair}. In addition to reflecting biased associations embedded in our %social discourse and 
historically recorded data, 
the blind application of these models turns biases in data into real-world discrimination against minority and marginalized social groups \citep{sap2019risk, zhao2017men}. This type of unintended discrimination has critical consequences in domains in which AI technologies are frequently used, such as education \citep{habib2019revolutionizing}, medicine and healthcare \citep{longoni2019resistance, briganti2020artificial}, and criminal justice \citep{rigano2019using}. One example of such biased models is Correctional Offender Management Profiling for Alternative Sanctions (COMPAS), a software for assessing defendants' risk of future crimes, widely used in courtrooms across the US. Recent efficacy assessments reveal that COMPAS assigns 77\% more likelihood of committing a future violent act to Black than White defendants regardless of their criminal history, recidivism, age, and gender \citep{angwin2016bias}.

AI models trained on biased datasets both \textit{reflect} and \textit{amplify} those biases. For example, the dominant practice in modern Natural Language Processing (NLP) --- which is to train AI systems on large corpora of social-media content, books, news articles, and other human-generated text data --- leads to models learning representational biases, such as implicit racial and gender biases \citep{caliskan2017semantics}. Models trained in this way are shown to  prefer European American names over African American names \citep{caliskan2017semantics}, associate
words with more negative sentiment with phrases referencing persons with disabilities \citep{hutchinson2020social}, make ethnic stereotypes by associating Hispanics with housekeepers, Asians with professors, and Whites with smiths \citep{garg2018word}, and assign men to computer programming and women to homemaking \citep{bolukbasi2016man}. 

Moreover, in NLP, language models are particularly susceptible to amplifying biases when their task involves evaluating language generated by or describing a social group \citep{blodgett2017racial}.
For example, previous research has shown that toxicity detection models associate documents containing features of African-American English with higher offensiveness than text without those features \citep{sap2019risk, davidson2019racial}. Similarly, \citet{dixon2018measuring} demonstrate that due to the high amount of hate speech on social media mentioning the word ``gay,'' models trained on social media posts are prone to classifying ``I am gay'' as hate speech. Applying such models for moderating social-media platforms can yield disproportionate removal of social media posts generated by or simply mentioning minority groups \citep{davidson2019racial, blodgett2020language}. This unfair assessment negatively impacts minority groups' representation in online platforms, which leads to disparate impacts on historically excluded groups \citep{feldman2015certifying}. 

Mitigating biases in hate speech detection, necessary for viable automated content moderation \citep{davidson2017automated, mozafari2020hate}, has recently gained momentum \citep[e.g.][]{davidson2019racial,dixon2018measuring,xia2020demoting,sap2019risk,kennedy2020contextualizing,prabhakaran2019perturbation} due to an increase in online hate speech and discriminatory language \citep{laub2019hate}, and the need for healthier online spaces, articulated by governments and the general public alike. Hate speech classification, like other NLP tasks, relies on data resources that potentially reflect real-world biases. Most current supervised algorithms for hate speech detection rely on (1) language models, which map textual data to their numeric representations in a semantic space; and (2) human annotations, which represent subjective judgements about the hate speech content of the text, constituting the training dataset. Both (1) and (2) can introduce biases into the final model. First, a classifier may become biased due to how the mapping of language to numeric representations is affected by stereotypical co-occurrences in the training data of the language model. For example, a semantic association between phrases referencing persons with disabilities and words with more negative sentiment in the language model can impact a classifier's evaluation of a sentence about disability \citep{hutchinson2020social}. Second, individual-level biases of annotators can impact the classifier in  stereotypical directions. For example, a piece of rhetoric about disability can be analyzed and labeled differently depending upon the social biases of the annotators. Although previous research has documented stereotypes in language models \citep{garg2018word, bolukbasi2016man, manzini2019black,swinger2019biases,charlesworth2021gender}, the impact of annotators' own biases on training data and models remains largely unknown. This gap in our understanding of the effect of human annotation on biased NLP models is the focus of this work.

In this paper, we investigate how hate speech annotation is impacted by social stereotypes and whether, as a result, hate speech classifiers trained on human annotations show human-like biases in their predictions. Study 1 investigates the role of people's social stereotypes in their hate speech annotation behavior in a large, nationally stratified sample in the US. Study 2 explores the impact of social stereotypes embedded in language on aggregated behavior of expert annotators in a large corpus of hate speech, which is used for training hate speech classifiers. Finally, Study 3 demonstrates how language-embedded stereotypes are associated with biased predictions in a neural network model for hate speech detection. Before presenting our studies, we briefly review the social psychological literature on social stereotyping and discuss the theoretical framework on which our analyses are based. 

\subsection{Social Stereotyping}

A comprehensive evaluation of human-like biases in hate speech classification needs to be grounded in social psychological theories of prejudice and stereotypes, in addition to how they are manifested in language \citep{blodgett2020language, kiritchenko2020confronting}. Stereotyping is a cognitive bias, deeply rooted in human nature \citep{cuddy2009stereotype} and omnipresent in everyday life. Stereotyping affords humans with the capacity to promptly assess whether an outgroup is a threat or not. Along with other cognitive biases, stereotyping impacts how individuals create their subjective social reality as a basis for social judgements and behaviors \citep{greifeneder2017social}. Stereotypes are often studied in terms of the associations that automatically influence judgement and behavior when relevant social categories are perceived \citep{greenwald1995implicit}. 

The Stereotype Content Model \citep[SCM;][]{fiske2002model} suggests that social perceptions and stereotyping form along two dimensions, namely \textit{warmth} (e.g., trustworthiness, friendliness, kindness) and \textit{competence} (e.g., capability, assertiveness, intelligence). To determine whether other people are threats or allies, individuals make prompt assessments about their warmth (good vs. ill intentions) and competence (ability vs. inability to act on intentions). The SCM's main tenet is that perceived warmth and competence underlie group stereotypes. Hence, different social groups can be positioned in different locations in this two-dimensional space, since much of the variance in stereotypes of groups is accounted for by these basic social psychological dimensions. For example, in the modern-day US, Christians and heterosexual people are perceived to be high on both warmth and competence, and people tend to express pride and admiration for these social groups. Asian people and rich people are stereotyped to be competent, but not warm. Elderly and disabled persons, on the other hand, are stereotyped to be warm, but not competent. Finally, homeless people, low-income Black people, and Arabs are stereotyped to be cold and incompetent \citep{fiske2007universal}. The warmth and competence dimensions appear to be universal features of social perception and predicted by the structural relations between groups \citep{fiske2002model}.

In three studies, we investigate (1) the general relationship between social stereotypes and hate speech annotation, (2) the particular relationship in curated datasets between social stereotypes and aggregated annotations of trained, expert annotators, and (3) social stereotypes as they manifest in the predictions of hate speech classifiers. We perform these studies with varying measures of stereotypes. In Study 1, we explore whether novice annotators' labeling behaviors are associated with their stereotypes of the target social groups.  In Study 2, we turn to  language-encoded stereotypes and the aggregated behavior of expert annotators. In Study 3, we then investigate whether the errors of classifiers, having been trained on annotations from experts, are due to the stereotypical bias against targeted social groups. In other words, we investigate whether the biases introduced by social stereotyping, either held by annotators or encoded in language, result in classifiers becoming biased along the corresponding stereotypical dimensions.

\section{Study 1}
\label{sec:novice}

%Cognitive biases, such as racial stereotypes, impact people's interpretation of language \citep[e.g.,][]{sap2019risk}. 
Here, we investigate the effect of individuals' social stereotypes on their hate speech annotations. Specifically, we aim to determine whether novice annotators' stereotypes of the mentioned social groups lead to higher rate of labeling text as hate speech. Furthermore, we aim to understand how stereotypes are associated with higher rates of disagreements among the annotators. 
We conduct a study on a nationally stratified sample  (in terms of age, ethnicity, gender, and political orientation) of US adults. First, we ask participants to rate eight US-relevant social groups on different stereotypical traits (e.g., friendliness). Then, participants are presented with social media posts mentioning the social groups and are asked to label the content of each post based on whether it attacks the dignity of that group. % In this context, we consider hate speech annotation a passive behavior, as it lacks any confrontational or explicit action toward the attacked group. Therefore, 
We expect the perceived warmth and/or competence of the social groups to be associated with participants' annotation behaviors, namely their rate of labeling text as hate speech and disagreeing with other annotators.

\subsection{Method}

\paragraph{Participants}
To achieve a diverse set of answers regarding the eight social groups under study, we selected a relatively large ($N$ = 1,228) set of participants. We recruited participants in a  US sample stratified across participants' gender, age, ethnicity, and political ideology through Qualtrics Panels. After filtering participants based on their performance on the quality-check items (described below), our final sample included 857 American adults (381 male, 476 female) ranging in age from 18 to 70 (\textit{M} = 46.7, \textit{SD} = 16.4), about half Democrats ($50.4\%$) and half Republicans ($49.6\%$), with diverse reported race/ethnicity ($67.8\%$ White or European-American, $17.5\%$ Black or African-American, $17.7\%$ Hispanic or  Latino/Latinx, $9.6\%$ Asian or Asian-American)\footnote{Please see the Supplementary Materials for analyses repeated with the full recruited sample}. 

\paragraph{Stimuli}
To compile a set of stimuli items for this study, we selected posts from the Gab Hate Corpus \citep[GHC;][]{kennedy2020gab}, which includes 27,665 social-media posts collected from the corpus of Gab.com \citep{pushshift_gab}, each annotated for their hate speech content by at least three expert annotators. We collected all posts with high disagreement among the GHC's (original) annotators (based on Equation \ref{equ:anno_dis} for quantifying \textit{item disagreement}) which mention at least one social group. 
We searched for posts mentioning one of the eight most frequently targeted social groups in the GHC: (1) women; (2) immigrants; (3) Muslims; (4) Jews; (5) communists; (6) liberals; (7) African-Americans; and (8) homosexuals. These social groups represent targeted US-related groups in GHC with regard to gender, nationality, religion, ethnicity, ideology, political affiliation, race, and sexual orientation. From the posts that satisfied this constraint, we selected seven posts per group, resulting in a set of 56 items in total.\footnote{Please see the Supplementary Materials for the list of all items.}

\paragraph{Explicit Stereotype Measure} 
We assessed participants' warmth and competence stereotypes of the 8 US social groups in our study based on their perceived traits for a typical member of each group. In other words, we asked participants to rate a typical member of each social group (e.g., Muslims) based on their ``friendliness'', ``helpfulness'', ``violence'', and ``intelligence''. Following previous studies of perceived stereotypes \citep{cuddy2007bias}, participants were asked to rate these traits from low (e.g., ``unfriendly'') to high (e.g., ``friendly'') using an 8-point semantic differential scale. We considered the average of the first three traits as the indicator of perceived warmth (Cronbach's $\alpha$s ranged between .90 [women] and .95 [Muslims]) and the fourth item as the perceived competence.  %Lastly, to analyze the impact of implicit biases of our pariticipants on their annotations judgments, each participant was assigned to complete an Implicit Association Tests \citep[IAT]{greenwald1998measuring}. 

\paragraph{Hate speech Annotation Task}
We asked participants to identify hate speech in the 56 items (described in the Stimuli section) after they were given a short definition of hate speech, provided by \citet{kennedy2020gab}:

\begin{quote}
    Language that intends to attack the dignity of a group of people, either through an incitement to violence, encouragement of the incitement to violence, or the incitement to hatred.
\end{quote}

We verified participants' understanding of the above definition of hate speech by explicitly asking them to verify whether they ``understand the provided definition of hate-based rhetoric.'' Participants could proceed with the study only after they  acknowledge understanding the definition. We then tested their understanding of the definition by placing three synthetic ``quality-check'' items among survey items, two of which included clear and explicit hateful language directly matching our definition and one item that was simply informational (see Supplementary Materials). Overall, 371 out of the original 1,228 participants failed to satisfy these conditions and their responses were removed from the data. The replication of our analyses using the original set of 1,228 participants is provided in the Supplementary Materials. 

%Moreover, we ask participants to provide their demographic information (race, sex, sexual identity, religious affiliation, religiosity, political orientation, race or ethnicity, education level, annual income, social class). 

\paragraph{Annotation Disagreement}
Throughout this paper, we assess annotation disagreement in different levels:
\begin{itemize}
    \item \textit{Item disagreement, $d_{(i)}$}: For an item $i$, we define item disagreement $d_{(i)}$ as the number of coder pairs that disagree on the item's label, divided by the number of all possible coder pairs. Specifically, if $n^{(i)}_{1}$ and $n^{(i)}_{0}$ show the number of hate and non-hate labels assigned to document $i$ respectively, item disagreement for document $i$ is calculates as
    \begin{equation}
    \label{equ:anno_dis}
        d_{(i)} = \frac{n^{(i)}_{1} \times n^{(i)}_{0}}{\binom{n^{(i)}_{1} + n^{(i)}_{0} }{2}}% = n^{pos}_i \times n^{neg}_i \times \frac{(n^{pos}_i + n^{neg}_i) \times (n^{pos}_i + n^{neg}_i  - 1)}{2}
    \end{equation}

    \item \textit{Participant item-level disagreement, $d_{(p, i)}$}: For each participant $p$ and each item $i$, we define participant item-level disagreement $d_{(p, i)}$ as the number of participants with whom $p$ agreed, divided by the total number of other participants who annotated the same item.  
    \item \textit{Group-level disagreement, $d_{(p, S)}$}: The group-level disagreement captures how much a participant disagrees with others over a specific set of items. For participant $p$ and a set of items $S$, we define the group-level disagreement $d_{(p, S)}$ by averaging $d_{(p, i)}$s for all items $i \in S$
    \begin{equation}
    \label{equ:group_dis}
        d_{(p, S)} = \frac{1}{|S|}\sum_{i \in S}d_{(p, i)}
    \end{equation} 
    
\end{itemize}

\paragraph{Statistical Analysis}

To explore participants' annotation behaviors relative to other participants, we rely on the Rasch model \citep{rasch1993probabilistic}. The Rasch model is a psychometric method that models participants' responses --- here, annotations --- to items by calculating two sets of parameters, namely the \textit{ability} of each participant and the \textit{difficulty} of each item.
To provide an estimation of these two sets of parameters, the Rasch model iteratively fine-tunes their values in order to fit the best probability model to participants' responses to items. Here, we use a Rasch model for each subset of items that mention a specific social group, leading to 8 ability scores for each participant.

It should be noted that while Rasch models consider each response as either correct or incorrect and generate an ability score for each participant, we assume no ``ground truth'' for the hate labels. Therefore, rather than interpreting the participants' score as their ability in predicting the correct answer, we interpret the scores as participants' \textit{tendency} for predicting higher number of hate labels. Throughout this study we use tendency to refer to the ability parameter, commonly used in Rasch models. Although items' difficulty scores are not assessed in our analyses, we can interpret them as items' \textit{controversiality}. 

\begin{figure}[t]
    \centering
    \includegraphics[width=.8\textwidth]{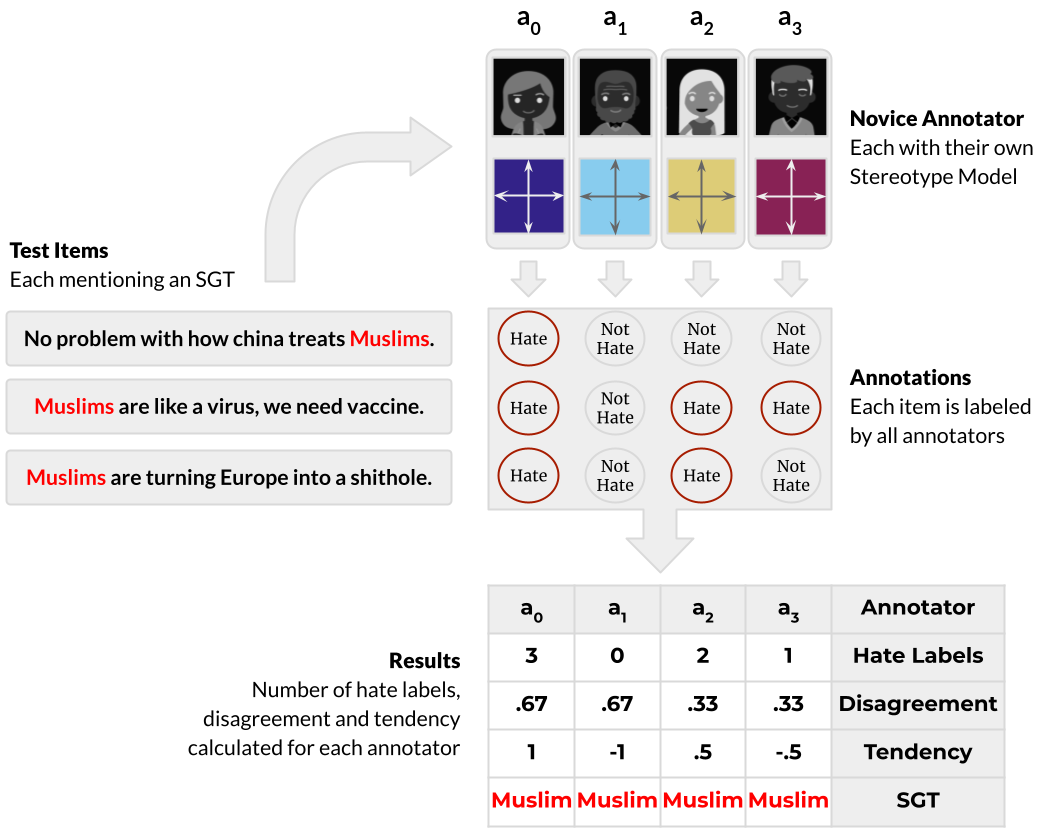}
    \caption{The overview of Study 1. Novice annotators are asked to label each social media post based on its hate speech content. Then, their annotation behaviors, per social group token, are taken to be the number of posts they labeled as hate speech, their disagreement with other annotators and their tendency to identify hate speech.}
    \label{fig:study1_overview}
\end{figure}

We estimate associations between participants' social stereotypes about each social group with their annotation behaviors evaluated on items mentioning that social group, namely, (1) the number of hate labels they detected, (2) their ratio of disagreement with other participants --- as quantified by \textit{group-level disagreement} ---
%the number of cases in which they disagreed with the majority vote, 
and (3) their tendency (via the Rasch model) to detect hate speech relative to others. To analyze annotation behaviors concerning each social group, we considered each pair of participants ($N$ = 857) and social groups ($n_{group}$ = 8) as an observation ($n_{total}$ = 6,856), which includes the social group's perceived warmth and competence based on the participant's answer to the explicit stereotype measure, as well as their annotation behaviors on items that mention that social group. We then fit cross-classified multi-level models to analyze the association of annotation behaviors with social stereotypes. Figure \ref{fig:study1_overview} illustrates our methodology in conducting Study 1.
All analyses were performed in \texttt{R} (3.6.1), and the \texttt{eRm} (1.0.1) package was used for the Rasch model.

\subsection{Results}

We first investigated the relation between participants' social stereotypes about each social group and the number of hate speech labels they assigned to items mentioning that group. The result of a cross-classified multi-level Poisson model, with the number of hate speech labels as the dependent variable and warmth and competence as independent variables, shows that a higher number of items are categorized as hate speech when participants perceive that social group as high on %higher implicit bias ($\beta=0.06, SE = 0.009, p $\textless .001) and 
competence ($\beta=$0.03, $SE =$ 0.006, $p $\textless .001). In other words, a one point increase in a participant's rating of a social group's competence (on the scale of 1 to 8) is associated with a 3.0\% increase in the number of hate labels they assigned to items mentioning that social group. However, warmth scores were not significantly associated with the number of hate-speech labels ($\beta=0.01, SE = 0.006, p = .128$).

We then analyzed participants' group-level disagreement for items that mention each social group. %When participants perceive higher implicit bias disagreed less with the majority votes based on the results of a Poisson model for predicting the disagreement ($\beta=-0.03, SE = 0.012, p $\textless .05), 
While the previous analysis uses a Poisson regression model to predict the \textit{number} of hate labels, here we use a logistic regression model to predict disagreement \textit{ratio} which is a value between 0 and 1.
The results of a cross-classified multi-level logistic regression, with group-level disagreement ratio as the dependent variable and warmth and competence as independent variables, show that 
participants disagreed more on items that mention a social group which they perceive as low on competence ($\beta=-0.29, SE = 0.001, p $\textless .001). In other words, a one point decrease in a participant's rating of a social group's competence (on the scale of 1 to 8) is associated with a 25.2\% increase in their odds of disagreement on items mentioning that social group. Warmth scores were not significantly associated with the odds of disagreement ($\beta=0.05, SE = 0.050, p = .322$).

Finally, we compared annotators' relative tendency to assign hate speech labels to items mentioning each social group, calculated by the Rasch models. As mentioned before, by tendency we refer to ability parameter calculated by Rasch model for each participant. We conducted a cross-classified multi-level linear model to predict participants' tendency as the dependent variable, and each social group's warmth and competence as independent variables. %implicit bias does not have a significant association with the \textit{ability} score ($\beta=0.04, SE = 0.031, p = .25$), however, 
The result shows that participants demonstrate higher tendency (to assign hate speech labels) on items that mention a social group they perceive as highly competent
($\beta=0.07, SE = 0.013, p $\textless .001). However, warmth scores were not significantly associated with participants' tendency scores ($\beta=0.02, SE = 0.014, p = 0.080$).

\begin{figure}[t]
     \centering
     \begin{subfigure}
         \centering
         \includegraphics[width=0.5\textwidth]{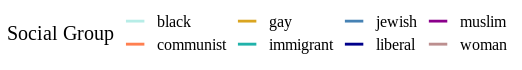}
         %\caption{Positive hate speech bias}
         \label{fig:hate_explicit}
     \end{subfigure}
     \hfill
     \begin{subfigure}
         \centering
         \includegraphics[width=\textwidth]{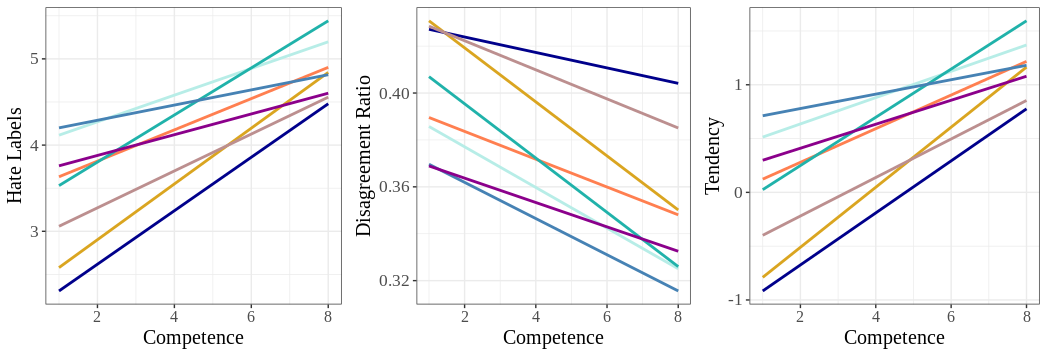}
         \label{fig:hate_implicit}
     \end{subfigure}
     \caption{The relationship between the stereotypical competence of social groups and (1) the number of hate labels annotators detected, (2) their ratio of disagreement with other participants -- as quantified by \textit{group-level disagreement}, and (3) their \textit{tendency} to detect hate speech -- as quantified by the Rasch model (left to right)}
     \label{fig:competence_novice}
\end{figure}

In summary, the results of Study 1 demonstrate that when novice annotators perceive a social group as high on competence they (1) assign more hate speech labels to, (2) disagree less with other annotators on, and (3) show higher tendency for identifying hate speech for documents mentioning those groups. Figure \ref{fig:competence_novice} represents these associations between annotation behaviors on our stimuli with regard to the stereotypical competence of the social group mentioned in each post.
These associations collectively denote that when annotators stereotypically perceive a social group as highly competent, they tend to become more sensitive or alert about hate speech directed toward that group. %In other words, novice annotators are more likely to label social media posts about stereotypically competent groups as hate speech and agree more on the final label of such posts. 
On the other hand, these annotators tend to be less sensitive about hate speech directed toward stereotypically incompetent groups. 
These results support the idea that hate speech annotation %, as a passive behavior, 
is affected by annotators' stereotypes against social groups, and specifically the perceived competence of the target social group.
\section{Study 2}
\label{sec:expert}

Hate speech annotation is marked by exceedingly high levels of inter-annotator disagreement \citep{ross2017measuring}, which can be attributed to numerous factors, including inconsistent definitions of hate speech, annotators' varying perception of the hateful language, or ambiguities of the text being annotated \citep{aroyo2019crowdsourcing}.
As Study 1 suggests, hate speech annotation is affected by annotators' stereotypes toward social groups mentioned in text. Specifically, the results suggest that, for novice annotators, stereotypically perceiving a social group as competent is associated with more sensitivity about and less disagreement on hate speech directed toward that group. Here, we explore the effect of social stereotypes, as encoded in language, on hate speech annotations performed by expert annotators in a large annotated dataset  of social media posts.  Contrary to novice annotators, we expect expert annotators to  focus more on groups that are the typical target of hate speech; namely, groups perceived as cold and incompetent. 
%engage in hate speech annotation because their training emphasizes on the very goals of the annotation process. %Perceived as an active behavior, 

Annotated datasets of hate speech rarely report psychological assessments of their annotators \citep{bender2018data,prabhakaran2021releasing}, and even if they do, little variance may exist among the few annotators who code an item. Therefore, rather than relying on annotators' self-reported social stereotypes, here we analyze stereotypes based on the semantic representation of social groups in pre-trained language models, which have been shown to reflect biases from large text corpora \citep{bender2021dangers}.

\begin{figure}
    \centering
    \includegraphics[width=\textwidth]{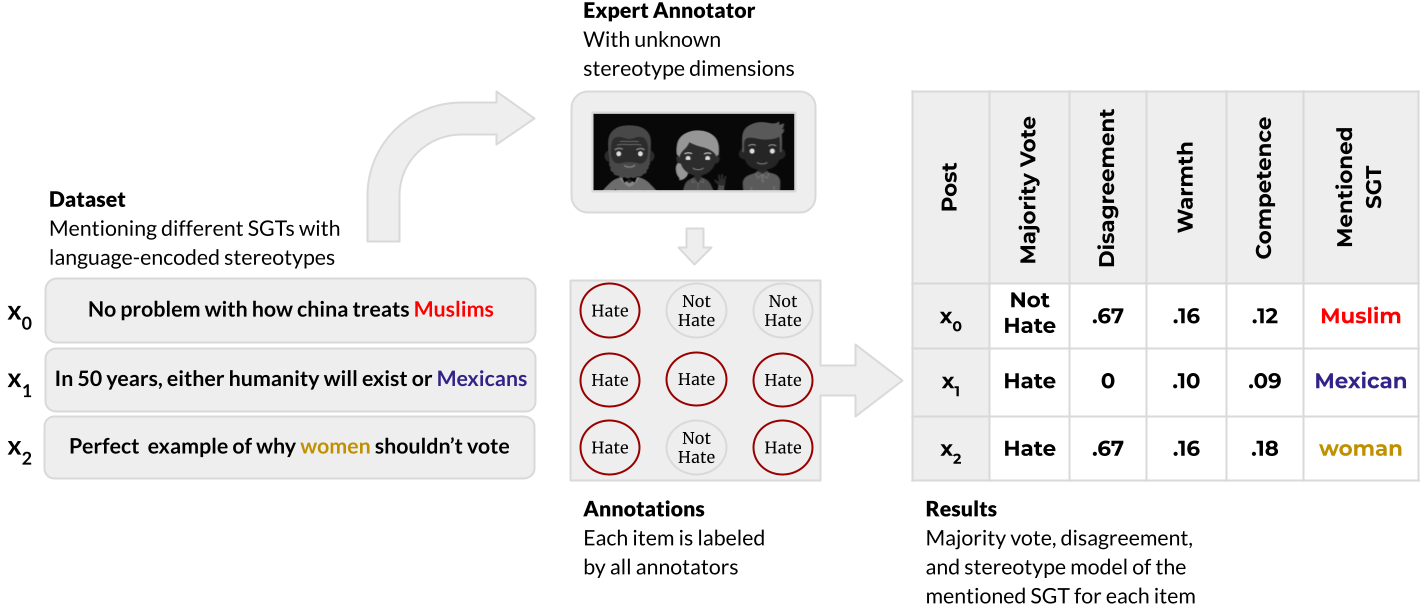}
    \caption{The overview of Study 2. We investigate a dataset of social media posts and evaluate the inter-annotator disagreement and majority label for each document in relation to language-encoded stereotypes of mentioned social groups. Contrary to Study 1, stereotypes do not vary at annotator-level.}
    \label{fig:study2_overview}
\end{figure}

We %first demonstrate the dependence of inter-annotator disagreement on the \textit{mentions} of social groups in text (see Supplementary Materials), and then 
propose that the activation of different social stereotypes upon observing social group tokens  leads to higher rates of disagreement. For example, annotators might disagree more when annotating a social media post that attacks ``Jews'' --- a stereotypically cold and competent minority group --- whereas they would agree on the label of a post that mentions ``Whites'' --- a stereotypically warm and competent group.
%We demonstrate that mere mention of social group tokens (\sgt s) are indeed associated with inter-annotator disagreement, despite the hate speech content of the text. 
We investigate this association using the social groups' stereotypes embedded in pre-trained language models and disagreement in hate labels. We ask, based on Study 1, whether lower tendency for, and higher disagreement on, hate speech directed toward stereotypically incompetent groups is tied to higher inter-annotator disagreement in a large dataset with high rates of hateful language. Figure \ref{fig:study2_overview} illustrates the methodology of Study 2, schematically.

\subsection{Method}

\paragraph{Data}
We used the GHC \citep[discussed in Study 1]{kennedy2020gab} which includes 27,665 social-media posts collected from the corpus of Gab.com \citep{pushshift_gab}, each annotated for hate speech content.
GHC includes hate speech annotations generated by 18 trained research assistants who completed in-person and thorough training based on a manual of hate-based rhetoric \citep{kennedy2020gab}. This dataset includes 91,967 annotations in total, where each post is annotated by at least three coders. On average, each annotator coded 5,109 posts ($Mdn = 4,044$). 
Fleiss's $\kappa$ for multiple annotators \citep{fleiss1971measuring} is 0.23,  and Prevalence-Adjusted and Bias-Adjusted Kappa \citep[PABAK;][]{byrt1993bias}, considered more appropriate for datasets with imbalanced labels such as the GHC, is 0.67. Based on our definition of item disagreement in Equation \ref{equ:anno_dis}, we computed the inter-annotator disagreement, and the majority vote for each of the 27,665 annotated posts and use them as the dependent variables in our analyses.
%The Fleiss's $\kappa$ agreement and PABAK scores were 0.23 and 0.67, respectively. %We calculate the annotation agreement for each post as the proportion of pairs of annotators who agreed with one another to all pairs of annotators.

\paragraph{Quantifying Social Stereotypes}
We collected a list of \sgt s by extending the list of identity terms suggested by \citet{dixon2018measuring}, obtaining synonyms via WordNet \citep{miller1995wordnet}. 
We then quantified social stereotypes directed toward each social groups in our list using the similarity of the semantic representation of each social group to the dictionaries of competence and warmth developed and validated by \citet{pietraszkiewicz2019big}. %Competence and warmth support systematic patterns of cognitive, emotional, and behavioral reactions, including ambivalent prejudices. %Past views of prejudice as a univalent antipathy have obscured the unique responses toward groups stereotyped as competent but not warm (e.g., rich) or warm but not competent (e.g., elderly) \citep{cuddy2008warmth}. 
The dictionaries were created following the Linguistic Inquiry and Word Count (LIWC) approach \citep{pennebaker2001linguistic} by a committee of social psychologists. These validated dictionaries have been shown to measure linguistic markers of competence and warmth reliably and efficiently in different contexts. The competence and warmth dictionaries consists of 192 and 184 tokens, respectively \citep{pietraszkiewicz2019big}. %Each item is either a token that represents a specific word (e.g., ``affinity'') or a token followed by a asterisk, that indicates the acceptance of all letters, hyphens, or numbers following its appearance (e.g., socia*).

\begin{figure}[t]
    \centering
    \includegraphics[width=.5\textwidth]{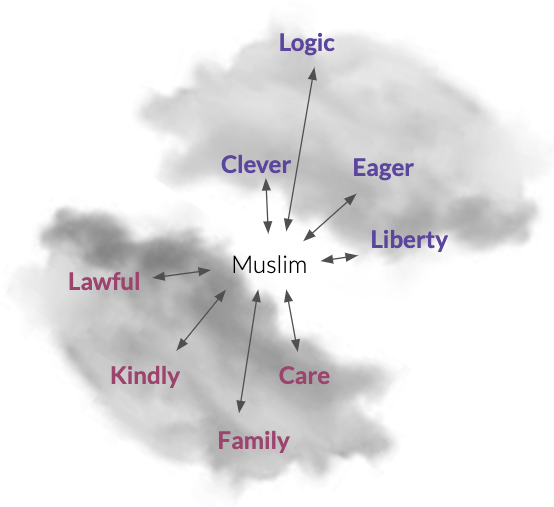}
    \caption{For each social group token, we averaged similarities of tokens from dictionaries of warmth (bottom-left) and competence (top-right), calculated in a semantic space of English words.}
    \label{fig:dict}
\end{figure}

Based on previous approaches for finding associations between words and dictionaries \citep{caliskan2017semantics, garg2018word, garten2018dictionaries}, we calculated the similarity of each \sgt~with the entirety of words in dictionaries of warmth and competence in a latent vector space.
Specifically, for each \sgt{}, $s$ and each word $w$ in the dictionaries  of warmth ($D_{w}$) or competence ($D_{c}$) we first obtain their numeric representation ($R(s)  \in \mathbb{R}^t$ and $R(w) \in \mathbb{R}^t$ respectively) from pre-trained English word embeddings \citep[GloVe;][]{pennington2014glove}. 
The representation function, $R()$, maps each word to a  $t$-dimensional vector, trained based on the word co-occurrences in a corpus of English Wikipedia articles. 
Then, the warmth and competence scores for each \sgt~were calculated by averaging the cosine similarity of the numeric representation of the \sgt~and the numeric representation of the words of the two dictionaries (Figure \ref{fig:dict}).
\begin{equation}
\label{equ:warmth}
    Warmth(s) = \frac{1}{|D_w|} \sum_{w \in D_{w}} cos(R(s), R(w))
\end{equation}

\begin{equation}
\label{equ:competence}
    Competence(s) = \frac{1}{|D_c|} \sum_{w \in D_{c}} cos(R(s), R(w))
\end{equation}

%, each as a value between $-1$ (completely dissimilar) and $+1$ (strongly similar). 

\begin{figure}[hbp]
     \centering
     \begin{subfigure}
         \centering
     \includegraphics[width=.8\textwidth]{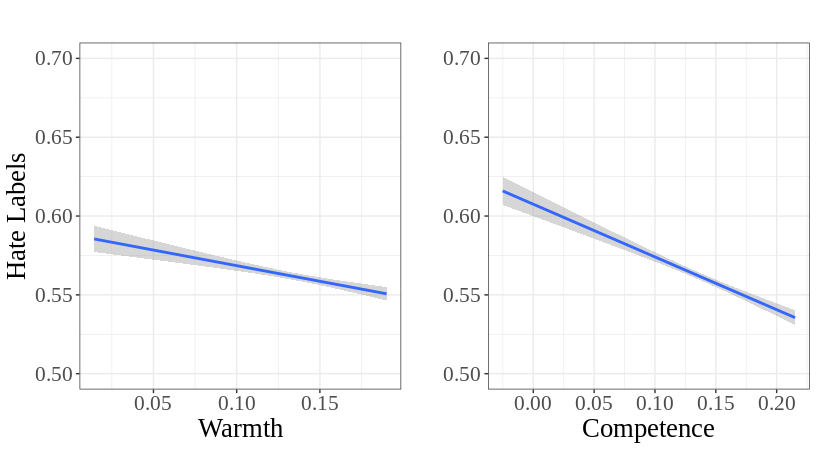}
     \end{subfigure}
     \hfill
     \begin{subfigure}
         \centering
         \includegraphics[width=.8\textwidth]{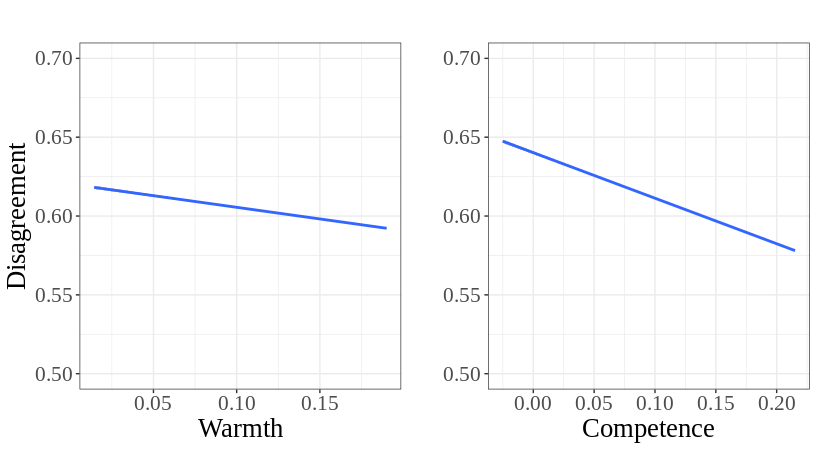}
     \end{subfigure}
     \caption{The effect of mentioned social groups' stereotype content, as measured based on their semantic similarity to the dictionaries of warmth and competence, on %expert annotators' performance. 
     annotators' disagreement.}
     \label{fig:stereo_expert}
\end{figure}

\subsection{Results}

We examined the effects of \sgt s on hate speech annotation based on their quantified social stereotypes. Specifically, we compared post-level  %hate speech labels and 
annotation disagreements with the mentioned social group's warmth and competence as measured by the cosine similarity of the vector representing the \sgt~and the vector representing the semantic notions of warmth and competence described earlier. For example, based on this method, ``man'' is the most  semantically similar \sgt~to the dictionary of competence ($Competence(man) = 0.22$), while ``elder'' is the \sgt~with the closest semantic representation to the dictionary of warmth ($Warmth(elder) = 0.19$). Of note, we investigated the effect of these stereotypes on hate speech annotation in social media posts that mention at least one \sgt~ ($N_{posts}$ = 5535).
Since some posts mention more than one \sgt, we considered each mentioned \sgt~as an observation ($N_{sgt}$ = 7550), and conducted a multi-level model, with mentioned \sgt s as the level-1 variable and posts as the level-2 variable.
We conducted two logistic regression analysis to assess the impact of (1) the warmth and (2) the competence of the mentioned social group as independent variables, and with the inter-annotator disagreement as the dependent variable. %,  scores were found to be predictive of agreement. %(Model's \textit{F}(3,49) = 3.30, \textit{p} = 0.03). 
The results of the two models demonstrate that both higher warmth ($\beta$ = -2.62, $SE$=0.76, \textit{p} \textless 0.001) and higher competence ($\beta$ = -5.27, $SE$ = 0.62, \textit{p} \textless 0.001) scores were associated with lower disagreement.
Similar multi-level logistic regressions with the majority hate label of the posts as the dependent variable and considering either social groups' warmth or competence as independent variables show that competence predicts lower hate ($\beta$ = -7.77, $SE$=3.47, $p = .025$), but there was no significant relationship between perceived warmth and the hate speech content ($\beta$ = -3.74, $SE$ = 4.05, \textit{p} = 0.355).

In this study, we demonstrated that language-encoded dimensions of stereotypes (i.e., warmth and competence)  are associated with annotator disagreement over the document's hate speech label. As in Study 1, annotators agreed more on their judgements about social media posts that mention stereotypically more competent groups. Moreover, we observed higher inter-annotator disagreement on social media posts that mentioned stereotypically cold social groups (Figure \ref{fig:stereo_expert}). 
While Study 1 demonstrated novice annotators' higher tendency for detecting hate speech targeting stereotypically competent groups, we found a lower likelihood of hate labels for posts that mention stereotypically competent social groups in this dataset. This discrepancy is potentially due to the fact that while novice annotators are more sensitive about hate speech directed toward stereotypically competent groups (e.g., Whites), these groups are not perceived as targets of hate by expert annotators' majority vote. %Therefore, expert annotators seem to focus on protecting groups that hate is frequently used against. %or they do not tend to be frequent targets of hate speech in the GHC. 

\section{Study 3} 
%Referencing a social group, as shown in Study 1, significantly provokes disagreements in hate speech annotation. Ultimately,

Previous research has demonstrated that NLP models, trained on human-annotated datasets, are prone to patterns of false predictions associated with specific social group tokens \citep{blodgett2017racial, davidson2019racial}. For example, trained hate speech classifiers may have a higher probability of assigning a hate speech label to a non-hateful post that mentions the word ``gay'' but are less likely to mislabel posts that mention the word ``straight.'' Such patterns of false predictions are known as prediction bias \citep{hardt2016equality, dixon2018measuring}, which impact models' performance on input data associated with specific social groups.
Previous research has investigated several sources leading to prediction bias, such as disparate representation of specific social groups in the training data and language  models, or the choice of research design and machine learning algorithm \citep{hovy2021five}.
However, to our knowledge, no study has evaluated prediction bias with regard to annotators' social stereotypes.
In Study 3, we investigate whether stereotypes about social groups influence hate speech classifiers' prediction bias toward those groups. 

\begin{figure}[t!]
    \centering
    \includegraphics[width=\textwidth]{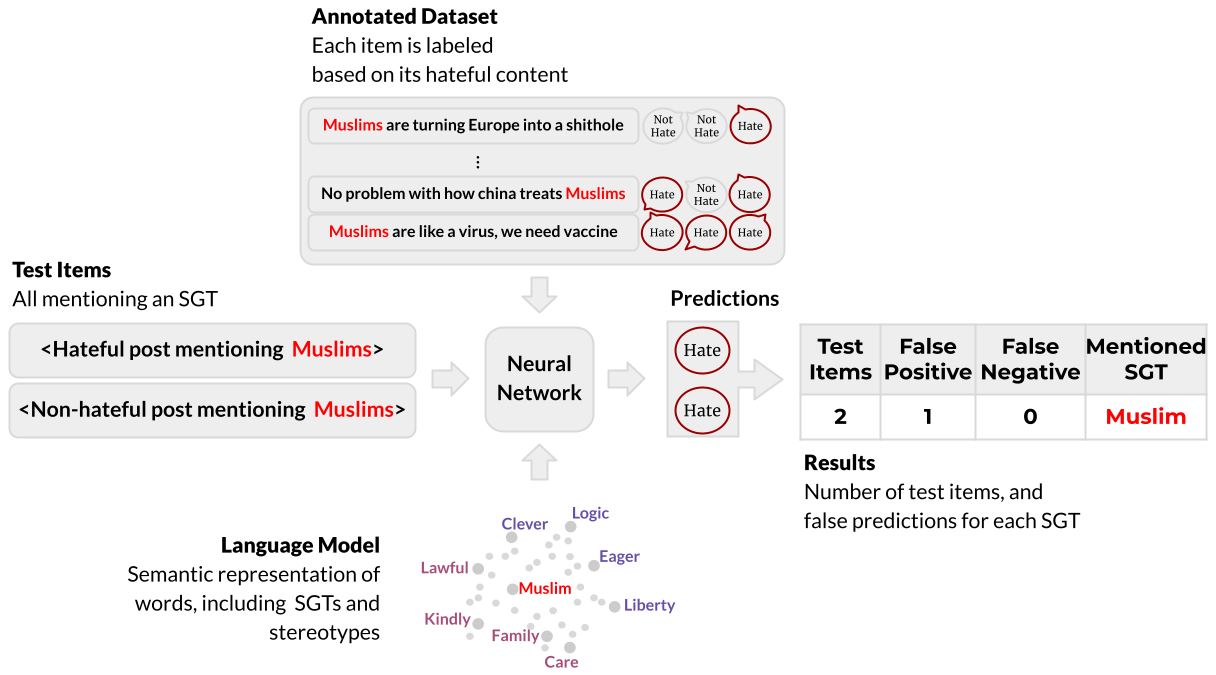}
    \caption{The overview of Study 3. During each iteration of model training the neural network learns to detect hate speech based on a subset of the annotated dataset and a pre-trained language model. The false predictions of the model are then calculated for each social group token mentioned in test items.}
    \label{fig:study3-overview}
\end{figure}

In Study 1, we examined social stereotypes reported by novice annotators to predict sources of variance in annotation behaviors. We discovered less disagreement and more annotator sensitivity about hate speech directed toward competent groups. In Study 2, we investigated the annotated dataset, created by aggregating expert annotators' judgements, and discovered high annotator disagreement and lower sensitivity about hate in documents mentioning social groups which are represented as cold and incompetent in language models.
%In Studies 1 and 2, we explored the effects of social group stereotyping on annotation behaviors and the annotated dataset. We examined social stereotypes either reported by annotators (Study 1) or embedded in natural language (Study 2) to predict sources of variance in annotation behaviors. We discovered more annotator reactivity about hate speech directed toward competent groups (Study 1) and high annotator disagreement in documents mentioning social groups which are stereotyped as cold and incompetent (Study 2). 
Accordingly, we expect hate speech classifiers, trained on the aggregated annotations, to be affected by such stereotypes, and perform less accurately and in a biased way on social-media posts that mention stereotypically cold and incompetent social groups. 
To detect patterns of false predictions for specific social groups (prediction bias), we first train neural network models on different subsets of an expert-annotated corpus of hate speech (GHC; described in Study 1). We then evaluate the frequency of false predictions provided for each social group and their association with the social groups' stereotypes. Figure \ref{fig:study3-overview} illustrates an overview of the methodology of this study.
%which respectively account for the incorrect labels assigned to hateful or non-hateful social-media posts, 
     
% \begin{figure}[ht]
%      \centering
     
%      \hfill
     
%         \caption{Social groups' (a) lower competence and (b) higher warmth are associated with higher biases in hate speech detection}
%         \label{fig:model-results}
% \end{figure}

\subsection{Method}

\paragraph{Hate Speech Classifier}
We designed a hate speech classifier based on a pre-trained language model, Bidirectional Encoder Representations from Transformers \citep[BERT;][]{delvim2019bert}. Given an input sentence, $x$, pre-trained BERT generates a multi-dimensional numeric representation of the text, $g(x) \in \mathbb{R}^{768}$. This representation vector is considered as the input for a fully connected function, \textit{h} which applies a softmax function on the linear transformation of $g(x)$: 

\begin{equation}
   h(g(x)) = softmax(W_h . g(x) + B_h) 
\end{equation}

During training, BERT's internal parameters and additional parameters, specifically $W_h \in \mathbb{R}^{768 \times 2}$ and $B_h \in \mathbb{R}^{2 \times 1}$, which are respectively the weights and bias matrices, are fine-tuned to achieve the best prediction on the GHC. We implemented the classification model using the \texttt{transformers} (v3.1) library of HuggingFace \citep{wolf2019huggingface}  and trained \textit{h} and \textit{g} in parallel during six epochs on an NVIDIA GeForce RTX 2080 SUPER GPU using the ``Adam'' optimizer \citep{kingma2014adam} with a learning rate of $10^{-7}$. %We perform this process a hundred times, each time selecting the training samples (80\% of the data) randomly and evaluating the predictions for the remaining samples (20\% of the data). %

To analyze the model's performance on documents mentioning each \sgt, we trained the model on a subset of the GHC and evaluated its predictions on the rest of the dataset. To account for possible variations in the resulting model, caused by selecting different subsets of the dataset for training, we performed 100 iterations of model training and evaluating. In each iteration, we trained the model on a randomly selected 80\% of the dataset ($n_{train} = 22,132$) and recorded the model predictions on the remaining 20\% of the samples ($n_{test} = 5,533$). Then, we explored model predictions ($n_{prediction} = 100 \times 5,533$), to capture false predictions for instances that mention at least one \sgt. By comparing the model prediction with the majority vote for that instance, provided in GHC, we detected all incorrect predictions. For each \sgt, we specifically capture the number of false-negative (hate speech instances which are labeled as non-hateful) and false-positive (non-hateful instances labeled as hate speech) predictions. For each \sgt~the false-positive and false-negative ratios are calculated by dividing the number of false prediction by the total number of posts mentioning the \sgt~.

\paragraph{Quantifying Social Stereotypes}
We quantified each social group's stereotype content based on Equations \ref{equ:warmth} and \ref{equ:competence} from Study 2. Recall that we calculated the similarity of each social group with dictionaries of warmth and competence, based on their semantic representations in a latent vector space of English. In each analysis, we considered either warmth or competence of social groups as the independent variable to predict false-positive and false-negative predictions as dependent variables.

\subsection{Results}

On average, the trained model achieved an $F_1$ accuracy of 48.22\% ($SD = 3\%$) on the test sets over the 100 iterations. Since the GHC includes a varying number of posts mentioning each \sgt, the predictions ($n_{prediction} = 553,300$) include a varying number of items for each \sgt \space (\textit{M} = 2,284.66, \textit{Mdn} = 797.50, \textit{SD} = 3,269.20). ``White'' as the most frequent \sgt~appears in 16,155 of the predictions and ``non-binary'' is the least frequent \sgt~with only 13 observations. We account for this imbalance by adding the log-transform of the number of test samples for each  \sgt\space as an offset to the regression analyses conducted in this section. 

The average false-positive ratio of \sgt s was 0.58 (\textit{SD} = 0.24), with a maximum of 1.00 false-positive ratio for several social groups, including ``bisexual'', and the minimum of 0.03 false-positive ratio for ``Buddhist.'' In other words, models always predicted incorrect hate speech labels for non-hateful social-media posts mentioning `bisexuals' while rarely making those mistakes for posts mentioning ``Buddhists''. The average false-negative ratio of \sgt s was 0.12 (\textit{SD} = 0.11), with a maximum of 0.49 false-negative ratio associated with ``homosexual'' and the minimum of 0.0 false-negative ratio several social groups including ``Latino.'' In other words, models predicted incorrect non-hateful labels for social-media post mentioning ``homosexuals'' while hardly making those mistakes for posts mentioning ``Latino''. These statistics are consistent with observations of previous findings \citep{davidson2017automated, kwok2013locate, dixon2018measuring, park2018reducing}, which identify false-positive errors as the more critical issue with hate speech classifiers.

\begin{figure}
         \centering
         \includegraphics[width=.95\textwidth]{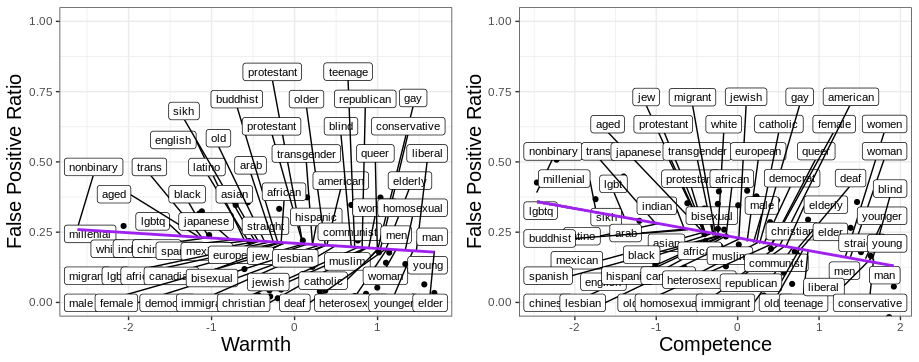}
         %\caption{Positive hate speech bias}
         \caption{Social groups' higher stereotypical competence and warmth is associated with higher false positive predictions in hate speech detection}
         \label{fig:fp}
\end{figure}

\begin{figure}
    \centering
    \includegraphics[width=.95\textwidth]{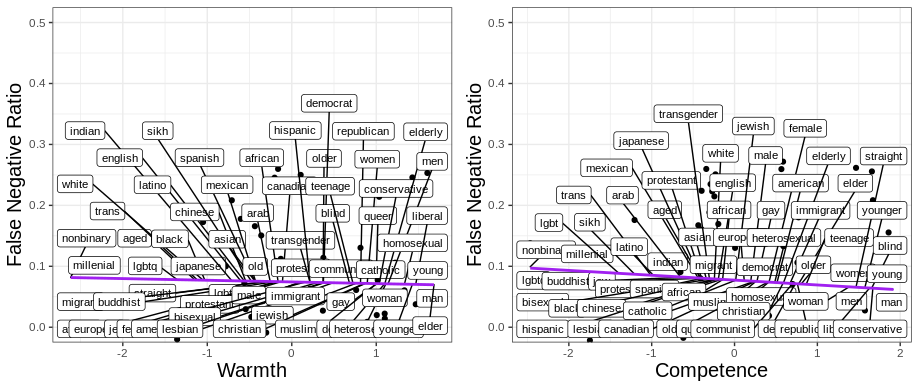}
    %\caption{Negative hate speech bias}
    \caption{Social groups' higher stereotypical competence and warmth is associated with higher false negative predictions in hate speech detection}
    \label{fig:fn}
\end{figure}

We conducted Poisson models to assess the number of false-positive and false-negative hate speech predictions for social-media posts that mention each social groups. In two Poisson models, false-positive predictions were considered as the dependent variable and social groups' (1) warmth or (2) competence, calculated from a pre-trained language model (see Study 2) were considered as the independent variable along with the log-transform of the number of test samples for each \sgt\space as the offset. The same settings were considered in two other Poisson models to assess false-negative predictions as the dependent variable, and either warmth or competence as the independent variable. The results indicate that the number of false-positive predictions is negatively associated with the social groups' language-embedded warmth ($\beta=-0.09, SE = 0.01, p $\textless .001) and competence scores ($\beta=-0.23, SE=0.01, p$\textless .001).
Therefore, texts that mentions social groups that are perceived as cold and incompetent are more likely to be misclassified as containing hate speech. In other words, a one point increase in the social groups warmth and competence is, respectively, associated with 8.4\% and 20.3\% decrease in model's false-positive error ratios. Moreover, the number of false-negative predictions is also negatively associated with the social groups' warmth ($\beta=-0.04, SE = 0.01, p $\textless .001) and competence scores ($\beta=-0.10, SE=0.01, p$\textless .001).
Therefore, texts that mention social groups that are perceived as cold and incompetent are more likely to be misclassified as not containing hate speech; one point increase in the social groups warmth is associated with 3.6\% decrease in model's false-positive error ratios and one point increase in competence is associated with 9.8\% decrease in model's false-positive error ratio. 
Figures \ref{fig:fp} and \ref{fig:fn} respectively depict the associations of the two stereotype dimensions with the proportions of false-positive and false-negative predictions for social groups.

In summary, this study demonstrates that hate speech classifiers trained on annotated datasets predict erroneous labels for documents mentioning specific social groups. Particularly, the results indicate that documents mentioning stereotypically colder and less competent social groups, which lead to higher disagreement among expert annotators based on Study 2, drive higher error rates in hate speech classifiers. This pattern of high false predictions (both false-positives and false-negatives) for social groups stereotyped as cold and incompetent implies that prediction bias in hate speech classifiers is associated with social stereotypes, and resembles human-like biases that we documented in the previous studies.
\section{General Discussion}

Social media is one of the most widely-used and powerful channels of communication across the globe, enabling us to express ourselves freely. Although free expression of opinions should be cherished, using online environments to attack, degrade, and call for violence upon other groups is an abuse of this digital public square. Accordingly, many social-media platforms such as Facebook and Twitter consider ``hate speech'' to be against their community policies and actively remove hateful content \citep{margetts2015political}. This removal of abusive language online is based on the notion that online hate speech is not an isolated social phenomenon and has been found to be correlated with real-world violence \citep[e.g.,][]{blake2021misogynistic}. Therefore, there has been an increasing interest in AI technologies, especially in NLP, to develop fair and accurate hate speech detection models by relying on human-annotated corpora and language models to curb online hate \citep{macavaney2019hate,ahmed2021tackling}. 

Here, we integrate theory-driven and data-driven approaches \citep{Wagner2021measuring} to investigate human annotators' social stereotypes as a source of bias in hate speech datasets and classifiers. %to assess fairness and transparency in classification models, demonstrate the cognitive bias sources for deleterious consequences of misclassification, and construct a more integrated and empirically informed psychological theory of hate speech annotation. 
In three studies, we combine social psychological theoretical frameworks and computational linguistic methods to make theory-driven predictions about hate-speech-annotation behavior and empirically test the sources of bias in hate speech classifiers.

In Study 1, we investigated the association between participants' self-reported social stereotypes against 8 different social groups, and their annotation behavior on a small subset of social-media posts about those social groups. We quantified participants' annotation behavior in terms of (a) the frequency and tendency of hate speech labeling, and (b) the extent of their disagreement with other novice annotators. Our findings indicate that for novice annotators judging social groups as competent is associated with a higher tendency toward detecting hate and lower disagreement with other annotators. We reasoned that novice annotators prioritize protecting the groups they perceive as warm and competent. These results can be interpreted based on the  Behaviors from Intergroup Affect and Stereotypes framework \citep[BIAS;][]{cuddy2007bias}: groups judged as competent elicit passive facilitation (i.e., obligatory association), whereas those judged as lacking competence elicit passive harm (i.e., ignoring). Here, novice annotators might tend to ``ignore'' social groups judged to be incompetent and not assign ``hate speech'' labels to inflammatory posts attacking these social groups. 

However, Study 1's results may not uncover the pattern of annotation biases in hate speech datasets labeled by expert annotators who are thoroughly trained for this specific task  \citep{patton2019annotating} and have specific experiences that affect their perception of online hate \citep{waseem2016you}.
In addition, expert annotation of hate speech is a goal-oriented behavior: expert annotators look for cues of derogation and dehumanization carefully to protect minoritized groups. In these cases, annotators actively evaluate not only label correctness but also the consequences of their labeling behavior. In Study 2, we examined the role of social group tokens and language-encoded stereotype content in expert annotators' disagreements in a large dataset containing outgroup-derogatory and dehumanizing language. %First, we found high levels of disagreement for texts that were ultimately labeled as hate speech by most annotators. Second, we found more disagreement in non-hate documents that contain social group tokens. In other words, the presence of a social group makes annotating a non-hateful text more controversial, as many annotators would disagree with the final label. Third, 
We found that, similar to Study 1,  %while the presence of a social group makes annotating a non-hateful text more controversial, texts mentioning groups that are stereotyped as cold and incompetent elicit more disagreement among expert annotators. However, 
texts that included groups that are stereotyped to be warm and competent (e.g., Whites) were highly agreed upon by annotators. However, unlike Study 1, we find that posts mentioning groups stereotyped as  incompetent --- typical targets of hate speech --- are more frequently labeled as hate speech. In simpler words, novice annotators tend to focus on protecting groups they perceive as competent, but expert annotators tend to focus on common targets of hate in the corpus.

To empirically demonstrate the effect of annotation bias on supervised models, in Study 3, we evaluated a hate speech classifier's performance an expert annotated dataset. %where we changed the SGT in each post with all possible SGTs in the existing dataset to evaluate model's reliance on specific SGTs. 
We used the count of false-positive and false-negative predictions to operationalize the classifier's unintended bias in assessing hate speech toward specific groups \citep{hardt2016equality}. Study 3's findings suggested that stereotype content of a mentioned social group is significantly associated with biased classification of hate speech such that more false-positive and false-negative predictions are generated for documents that mention groups that are stereotyped to be cold and incompetent. These results demonstrate that biased predictions are more frequent for the same social groups that evoked more disagreements between annotators in Study 2. Similar to \citet{davani2021dealing}, these findings specifically challenge supervised learning approaches that only consider the majority vote for training a hate speech classifier and dispose of the annotation biases reflected in inter-annotator disagreements. 
%The results of Study 2 and 3 collectively show that cognitive biases lead to annotation disagreement which is further reflected in biased performance of supervised classifiers. 
%The highest rate of false-positives belonged to ``Muslim'' and ``Jew'', meaning that if other SGTs get replaced by these two terms, they will become substantially more likely to be detected by the algorithm as containing hateful rhetoric. 

Our work has implications for training hate speech classifiers---and more broadly AI and NLP models---because of the legitimate concern that these technologies may perpetuate social stereotypes and societal inequalities \citep{barocas2016big}. Our findings suggest that hate speech classifiers trained on human annotations will also acquire particular social stereotypes toward historically marginalized groups. Our results have three specific and direct implications: First, one way to decrease unintended bias in these algorithms is to diversify expert annotation teams, so that annotators come from different ideological and social backgrounds; by doing so, coders may not agree on a label to (over)protect an ingroup or ``ally'' groups (leading to false positives), or to ignore actually hateful content targeting ``blameworthy'' groups (leading to false negatives). Second, supervised learning approaches may benefit from modeling annotation biases, which are reflected in inter-annotator disagreements, rather than the current practice, which is to treat them as unexplained noise in human judgement, to be disposed of through annotation aggregation (e.g., majority voting). The third implication of the present work concerns social science theory development: while societal contexts have been often considered in social science theories, existing theoretical frameworks were not advanced with the vast societal reach of AI in mind. Our work is an example of how existing theories can be applied to explain the novel interactions between algorithms and people. Large amounts of data that are being constantly recorded in ever-changing socio-technical environments call for integrating novel technologies and their associated problems in the process of theory development \citep{muthukrishna2019problem,bail2014cultural,Wagner2021measuring}.

\subsection{Conclusion} 
%In sum, our work contributes to an increased understanding of the evolution of social media platforms and their underlying psychology invites an integration of social science theories and AI systems that automatically regulate or otherwise impacts human interactions. 
Our work suggests that societal stereotypes are associated with annotators' hate-speech-annotation behavior. Moreover, bias in hate speech detection models is in part due to annotation disagreements rooted in annotators' own biases, namely perceptions of different social groups' warmth and competence. Our framework, combining social psychological and computational linguistic methods, makes novel predictions to understand the effect of human biases on natural language processing models. 

Our findings come at a time when people's attitudes are largely shaped by online social environments \citep{cinelli2021echo}, where online political debates are becoming more contentious and polarized \citep{carpenter2020political}, and where social media companies such as Facebook and Twitter are playing an influential role in socio-political discourse \citep{finkel2020political}. The timeliness of the issue, given the current social milieu, highlights how not attending to human-like biases of AI systems may impact various socio-psychological aspects of people's lives both online and offline.

\bibliographystyle{apa}
\bibliography{bib}
\newpage
\setcounter{page}{1}

\section{Supplementary Materials}
\subsection{Study1}

\paragraph{Test and Annotation Items}
Table \ref{tab:test_items} and \ref{tab:all_items} respectively represent the 3 test items and all annotation items labeled by annotators in Study 1.

\begin{table}[]
    \centering
    \begin{tabular}{lc}
    \hline
        Test Item & Expected Label \\\hline
        I want to see Sandirians raped to death collectively. & Hate \\
        Now if we could only get a few jew heads on pikes that would be great. & Hate \\
        Over 99\% of Syrian refugees admitted to UK in 2017 were Muslims. & Non-Hate \\\hline\\
    \end{tabular}
    \caption{Test items in annotation survey, participants were filtered based in their correct answers to this items.}
    \label{tab:test_items}
\end{table}

\begin{longtable}
    {p{.75\textwidth}p{.15\textwidth}}
    \hline
        Annotation Item & Target \\\hline
        Texans in the vast plains, rivers, and the farmlands must gather, arm up, and declare independence from the United States of America. We are a nation of warriors that fell at the Alamo only to rise victorious, just as the Spartans did in the narrow pass against the Persians. The USA needs us. We do not need those sickening f[**]ktards. & Immigrant \\
        So if robots can plant the food, pick the food, and serve the food, why do we need illegal aliens again? Not to vote democrat. They have already moved on to simply changing the counts.& Immigrant \\
        The Kalergi Plan for White Genocide, (funded by George Soros), through unending Turd World Mass Migration into ALL White Nations. They Come. They breed, rape, terrorize, defecate, leech, infect and plunder... WE DIE!!!!& Immigrant \\
        Lesson: be careful what you wish for. Especially if you are an idiot. Never could figure out how people with such a better work ethic come from such shit hole countries? They are like Superman. Get near their home countries \& they become worthless, but super worker here? Strange how that happens. Like Work kryptonite.& Immigrant \\
        Go back to your shithole and learn how to be civilized.& Immigrant \\
        Everything on Instagram is in Portuguese now! Build that Wall!& Immigrant \\
        You know your country is MULTICULTURAL... When there's armed patrols on the streets.& Immigrant \\
        A church in Berlin was damaged after unknown attackers pelted stones at the building while shouting ``Allahu Akbar!'' & Muslim\\
        ``A place to discuss Jesus with fellow clean-living Christians''. lol, that description. But wasn't Christianity responsible for the church pedophilia?& Muslim\\
        I have no problem with how China treats these bastard Uighurs. They are only treating them like how the rest of us get treated by muzzies.& Muslim\\
        Look forward to reading UR next book. My 1st thought is that we are in this situation through ``cock-up'' rather than ``conspiracy''. Or more precisely, ``cock-up'' on our side \& something that's not quite conspiracy on the Muslim side: them just acting as they always do. We should have learned from History, but we were complacent.& Muslim\\
        Muslims are like a virus. We need the vaccine NOW.& Muslim\\
        Islam doesn't seek to assimilate into ANY culture, it seeks to assimilate cultures into it, you're naive to believe any differently with 1400 years of evidence.& Muslim\\
        Five alleged Muslim extremists accused of training children to carry out school shootings were released on bond after their attorney complained that the men were only being targeted because they’re ``black and Muslim,'' as opposed to ``white and Christian.''& Muslim\\
        Commies, They need to be driven from this world. Now. And forever. Misanthrope! & Communist\\
        There are two types of socialists: the evil socialist, and the stupid socialist. Evil socialists are despots who seek control. Stupid socialists think socialism works and is good.& Communist\\
        Tulsi Gabbard is basically an out-and-out socialist and open-borders zealot. That's not crazy enough for modern D's, though. You have to also want war with Russia. And the poor girl just couldn't get her mind right on that issue.& Communist\\
        This is Antifa, folks. Filthy, degenerate communist subhumans. Every single one.& Communist\\
        Who cares what John Dean says. The commies keep living in the past.& Communist\\
        IF the libtards and socialists (also known as libtards) keep trying to bring down OUR President, me and 63,000,000 of my closest friends will be in your face BIGGLY. And we know what to do because I am Jungle Warfare qualified and I did bring home 3 Bronze Stars from the Nam... AND I have never forgotten my oath.& Communist\\
        Yesterday, Judge Naomi B ruled that Trump can't block commies/animals from his private Twit acct because 1st Amend.& Communist\\
        ``Top psychiatrist: today’s left are mentally ill''. We already knew this, but it's nice to have it officially confirmed.& Liberal\\
        Now that Infowars has been hit by the Silicon Valley, I am amazed Twitter didn’t join in. But they are the same liberal far left trash and probably wanted to join the other kids kicking a dog when it’s down. Scum.& Liberal\\
        Bitching on social media may be cathartic, but the best way to fight the biased lefty press is not to watch, mention, click, link, post, or refer to their lefty crap. Even by complaining (mentioning), you’re giving them data to sell to their advertisers. Take away their ability to sell your data, take away your eyeballs and they’ll go out of business over time.& Liberal\\
        Popular leftist terms that do not exist: - Toxic Masculinity - Patriarchy - White Privilege. No matter how long the gender studies majors at McDonald's try to push this propaganda we must push back with facts.& Liberal\\
        ``I find wearing these little yellow stars on the chest is a good way to expiate our past sins, it reminds us of our privilege and keeps the memory of our crimes alive now that the last nazi is dead.'' Conservative, circa 2025.& Liberal\\
        Liberals: These bikini coffee chains are sexist and a bad influence on impressionable young girls. Also Liberals: Let’s take our kids down to the Folsom street fair where they can see transvestites swinging their junk around and getting whipped in public.& Liberal\\
        That hair probably costs over \$200 including the trip to the salon, and she's holding an \$800 iPhone... She's wearing makeup and jewelry, too, while she howls \& cries about how oppressed she is. Liberals have no sense of reality.& Liberal\\
        He did fuck all except promoting violence against white people, giving money to terrorism and trying and starting a race war... Useless President who still comes up first on auto-correct after typing President...& African-American\\
        Black Conservatives on Twitter: CNN's @JeffreyToobin called Antifa, a left-wing political terror group with few, if any, black participants, an ``African-American organization.'' These violent stereotypes are racist and wrong. Using it to smear POTUS Trump — even worse!& African-American\\
        Planned Parenthood is fighting back against Blacks catching on to what Margaret Sanger was all about: Limiting the Black population. It's why she formed PP in the first place. So now, billboards making it look bright and shiny and liberating to kill off their unborn.& African-American\\
        Guardian journalist, Daniel José Camacho, spews anti-white hatred after Super Bowl. (Guardianista nonwhite journalists openly now show their anti-white racism).& African-American\\
        It doesn’t matter if there are non-whites who share some of our nationalist views. they are still different from us. Should they be allowed in our ethnostate? If it were up to me, the answer is no. I don’t support a policy of, ``anyone who’s like Joe the Jew is welcome in our white ethnostate.'' The whole idea of an ethnostate is to preserve our genetics and identity.& African-American\\
        Mixing cultures is an issue, you end up with a mutt race.& African-American\\
        Western world is built on RACISM -- BBC Newsnight guest makes SHOCK announcement. If not for the white man this ``professor of black studies'', yeah I know LOL, would be walking naked amongst the lions \& zebras in Apefreaka.& African-American\\
        Horrific! A friend and former business partner who also is a retired policeman sent this along to me today. It describes a case he had, which was the wreckage left behind by one homosexual priest. Predators. Gay predators. Which sounds like a joke.& Gay\\
        People who say ``nobody's born a racist, you have to be taught'', are also the ones who tell their children to be gay degenerates and killing white people is right.& Gay\\
        Romney is remembered for the Obamacare prototype, for not fighting Gay Marriage, Speaking in front of Planned Parenthood, Being part of the Vulture Capital firm Bain, that destroyed companies and US Jobs with debt, and showing contempt for both the poor and the Deplorables. & Gay\\
        When even a flag triggers them, then you know that Hollywood is full of butthurt snowflakes!& Gay\\
        And here we see the fa***t it its natural habitat& Gay\\
        I said homosexuality is NOT natural and have been suspended until I take down my tweets, they will have a long wait, I hardly use twitter.& Gay.\\
        Pornography is a filthy tool by ((them)) to normalize miscegenation, homosexuality, and other degenerate perversions.& Gay \\
        Well, yea, but they also believe the greeks were Israelites...& Jew\\
        If they were only work camps that got twisted later on into death camps, does that mean the jew sees being made to finally work and not be a leech, as the *same* as death?& Jew\\
        In 50 years either humanity will exist or Israel will exist. Not both.& Jew\\
        Why do you follow Jew hater Cantwell? Oh, don't tell me your friend Danny is Jewish. So was Soros.& Jew\\
        Trump is the king of neo nazis and I am a true neo nazi. You're a fa***t jew boomer.& Jew\\
        Saying that the Holocaust happened makes you an Anti-Semite because it means you wanted 6 million jews to die.& Jew\\
        Why do people here believe the Holocaust happened? Why would you WANT TO BELIEVE SUCH AN ATROCITY HAPPENED TO THE JEWS? Of course it didn't happen, otherwise the world wouldn't have Israel and all the Nobel prize winners that came after WW2.& Jew\\
        Sharia May is to set up a new unit to deal with fake news. Stupid woman, if she wants to stop fake news, just shut down the BBC.& Woman\\
        Feminism leads to a decrease in the birth population which then requires mass migration from the 3rd world to support the aging population since feminists aren't having enough children to replace the population.& Woman\\
        Feminism is an infection!& Woman\\
        Damn that bitch is dumb.. with her pink hair, fucking Garth glasses and shitty old novelty guitar.& Woman\\
        Front Hole', Please explain the objectification of women in the current context. Thanks.& Woman\\
        Perfect example of why women shouldn't be voting.& Woman\\
        Feminism is now Female Supremacism as women have all the rights that men have and they still have special privileges in the courts and in our culture. Cultural Marxists who hate America, Whites, and the West are the leaders and preachers of Third Wave Feminism.& Woman\\\hline \\
    \caption{All annotated items in Study 1: for each social group understudy, 7 social media posts mentioning that social group is considered in the study.}
    \label{tab:all_items}
\end{longtable}

\paragraph{Study of All Annotators}
We replicate the results of Study 1, on the whole set of participants ($N$ = 1,228).
We first investigated the relation between participants' social stereotypes about each social group and the number of hate speech labels they assigned to items mentioning that group. The result of a cross-classified multi-level Poisson model, with number of hate speech labels as the dependent variable and warmth and competence as independent variables, shows that a higher number of items are categorized as hate speech when participants perceive that social group as high on 
competence ($\beta=$0.02, $SE =$ 0.005, $p $\textless .001). In other words, one point increase in a participant's rating of a social group's competence (on the scale of 1 to 8) is associated with a 1.9\% increase in the number of hate labels they assigned to items mentioning that social group. However, warmth scores were not significantly associated with the number of hate-speech labels ($\beta=0.01, SE = 0.006, p = .286$).

We then analyzed participants' group-level disagreement for items that mention each social group. 
The results of a cross-classified multi-level logistic regression, with group-level disagreement ratio as the dependent variable and warmth and competence as independent variables, show that 
participants disagreed more on items that mention a social group which they perceive as low on competence ($\beta=-0.17, SE = 0.034, p $\textless .001). In other words, one point decrease in a participant's rating of a social group's competence (on the scale of 1 to 8) is associated with a 15.5\% increase in the odds of disagreement on items mentioning that social group. Contrary to the original results, warmth scores were also significantly associated with the odds of disagreement ($\beta=0.07, SE = 0.036, p = .044$).

Finally, we compared annotators' relative \textit{tendency} to assign hate speech labels to items mentioning each social group, calculated by the Rasch models. As mentioned before, by \textit{tendency} we refer to \textit{ability} parameter calculated by Rasch model for each participant. We conducted a cross-classified multi-level linear model to predict participants' \textit{tendency} as the dependent variable, and each social group's warmth and competence as independent variables. %implicit bias does not have a significant association with the \textit{ability} score ($\beta=0.04, SE = 0.031, p = .25$), however, 
The result shows that participants demonstrate higher \textit{tendency} (to assign hate speech labels) on items that mention a social group they perceive as highly competent
($\beta=0.04, SE = 0.010, p $\textless .001). Warmth scores were only marginally associated with participants' \textit{tendency} scores ($\beta=0.02, SE = 0.010, p = 0.098$).
Except for the significant association of warmth stereotype with the odds of disagreement, other results are the same for both analyses. 

\paragraph{Implicit Bias}
To analyze the impact of implicit biases of our pariticipants on their annotations judgments, each participant was assigned to complete an Implicit Association Tests \citep[IAT]{greenwald1998measuring}. Each IAT assesses participants' implicit bias toward one of the 8 groups studied on Study 1 (\~150 participants for each IAT)
The IAT \citep{greenwald1998measuring} is a computer-based task designed to measure the strength of automatic associations between two opposing target categories and two opposing attributes.
In each trial, participants are instructed to categorize a stimulus (e.g., a word) as quickly and accurately as possible into one of two target categories or two attributes. In a first combined block, the two target categories and the two attributes are located with a certain associative pattern. In a second combined block, the location of the target categories is switched. A measure of the implicit associations can be obtained computing the difference between the mean latencies of the first and the second combined block. Previous work has shown that IAT scores can be used to assess attitudes and stereotypes, showing adequate levels of criterion validity \citep{greenwald2009understanding} and less proneness to impression management concerns compared with self-report measures \citep{vecchione2014fakability}.

In our study, the IAT provided a mechanism to quantify the participants' bias in preferring a group over a randomly selected pair. The social group pairs assessed in the IAT were the social groups mentioned in the hate speech items (Mexican vs American, Christianity vs Islam, Communism vs Capitalism, Liberal vs Conservative, White vs Black, Gay vs Straight, Female vs Male and Jewish vs Christian). As the result, the calculated IAT score for each participants represents their implicit bias (a value between -1.5 and +1.5) for a specific social group, such that a positive value would represent a higher bias against that group.

We investigated the relation between participants' implicit bias about each social group and the number of hate speech labels they assigned to items mentioning that group. We conducted a multi-level Poisson model, with social group as the level-1 variables, number of hate speech labels as the dependent variable and the implicit bias score as independent variables. The result shows that a higher number of items mentioning a social group are categorized as hate speech when participants show higher implicit bias for that group ($\beta=$0.09, $SE =$ 0.027, $p $\textless .01). In other words, one point increase in a participant's implicit bias score for a social group is associated with a 0.1\% increase in the number of hate labels they assigned to items mentioning that social group.

\subsection{Study 2}
First, in order to show that annotating hateful rhetoric leads to high levels of disagreement, even among expert annotators, we compared the occurrence of inter-annotator disagreement between hateful and non-hateful social-media posts.
Two-sample permutation tests (5,000 permutations) based on mean and median suggest that annotators disagree more on the posts which are labeled as hate speech ($M = 0.50$, $Md = 0.67$, $SD = 0.28$) compared with those that are labeled as not hateful ($M = 0.13$, $Md = 0.00$, $SD = .26$) by the majority vote ($p$ \textless .001). %The average disagreement score of hateful posts is calculated as 0.50 ($SD = 0.28$) compared with average disagreement of 0.13 ($SD = 0.26$) for content not labeled as hate speech. 
This effect is not surprising since hate speech annotation is shown to be a non-trivial task that requires careful consideration of the social dynamics between who generates a piece of text and who perceives the text. In addition, recognizing non-hate content is substantially easier than flagging inflammatory and \textit{potentially} hateful content \citep{waseem2016hateful}. 

We then assessed the association of textual mentions of social groups with inter-annotator \textit{item disagreements}. Two-sample permutation tests (5,000 permutations) based on mean and median suggest that posts which mention \sgt s triggered more disagreement ($p$ \textless .001) such that, in presence of \sgt s the averaged item disagreement is 0.30 ($Md = 0.00$, $SD = 0.32$), as opposed to averaged item disagreement of 0.13 ($Md = 0.00$, $SD = 0.26$) for posts without \sgt s. 

\begin{figure}
    \centering
    \includegraphics[width=.5\linewidth]{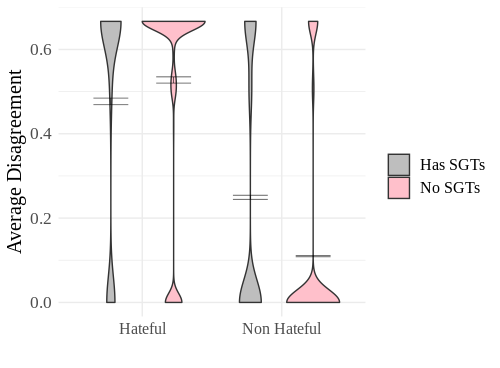}
    \caption{Disagreement scores on different subsets of the dataset, based on whether the posts include hate speech and social group tokens (SGT). The horizontal lines demonstrate the error bars.}
    \label{fig:expert_stat}
\end{figure}

To examine the interaction of presence of \sgt s and hate speech label, we conducted a two-way ANOVA to compare item disagreements, considering the binary hate label along with the presence of \sgt s as the factors. Both factors (presence of \sgt~and hate speech content) are significantly associated with item disagreement ($p$ \textless .001 for both). Figure \ref{fig:expert_stat} shows the distribution of disagreement scores for different subsets of the dataset. 
Generally, mentioning an \sgt~leads to higher inter-annotator disagreement, however, in hateful posts, mentioning an \sgt~actually results in annotators agreeing.  

\end{document}